# Artificial Intelligence-Assistant Cardiotocography: Unified Model for Signal Reconstruction, Fetal Heart Rate Analysis, and Variability Assessment

**Xiaohua Wang[1,2,*], Kai Yu[2], XuXiao Liang[2], Liang Wang[2], Chao Han[2]**

[1]Artificial Intelligence Research Center, Bengbu Medical University, Bengbu 233000, China.
[2]CHARMMIRAEL Biotech Co., Ltd, Nanjing 210000, China.
[*]E-mail: virgo_wang@msn.com

## Abstract

The monitoring of fetal heart rate (FHR) and the assessment of its variability are crucial for preventing fetal compromise and adverse outcomes. However, traditional methods encounter limitations arising from equipment performance, data transmission, and subjective assessments by doctors. We have developed a tailored AI-based FHrCTG model specifically for FHR monitoring, which effectively mitigates noise interference and precisely reconstructs signals. Our model was pre-trained on a massive dataset consisting of 558,412 unlabeled data points and further refined using 7,266 expert-reviewed entries. To validate FHR, we introduced the Intersection Overlapping Labels (IOL) approach, which transforms rate analysis into categorical judgments. Testing revealed that our model demonstrates high sensitivity and specificity in detecting critical FHR decelerations (89.13% and 87.78%, respectively) and accelerations (62.5% and 92.04%, respectively). Furthermore, based on Fischer's criteria for clinical application, our model achieved impressive AUC scores of 0.7214 and 0.9643 for verifying FHR periodicity and amplitude variation, respectively.

## Introduction

In the context of diverse racial and cultural backgrounds, the arrival of new life is always regarded as a paramount event. However, pregnancy, as a lengthy and delicate phase, can lead to serious or irreversible health consequences with any negligence[1]. During this period, fetal monitoring holds a pivotal position and constitutes a complex, meticulous task.

Hypoxic environments are one of the crucial factors that exert profound impacts on fetuses[2,3]. As fetuses rely entirely on the utero-placental blood flow in the umbilical cord for oxygen supply, the continuity and stability of this blood flow are of utmost importance. Any disruption or sudden event can lead to a decrease in the oxygen concentration delivered to the fetal arterial blood[4]. Such hypoxic conditions can have severe long-term effects on infants, including brain damage, developmental delays, and, in extreme and potentially fatal cases, infant death[5].

This situation is not without manifestation. When the fetal oxygen level in the mother's body is low, its sympathetic and parasympathetic nervous systems attempt to compensate by altering cardiac output and prioritizing blood supply to vital organs such as the heart and brain[6,7]. These manifestations are often associated with changes in the FHR[2-4]. Therefore, in the routine monitoring of fetuses, it is imperative for obstetricians with diagnostic and therapeutic expertise to frequently use instruments to monitor FHR and, based on their experience, identify fluctuations to prevent adverse outcomes for the infant[6-8]. Nonetheless, in practical operation, the sensitivity of clinical detection of fetal compromise remains relatively low, potentially leading to delayed delivery of at-risk infants and subsequently increasing the risk of adverse pregnancy outcomes[9].

The FHR inherently embodies a intricate amalgamation of diverse noise sources, encompassing a spectrum of noise signals such as maternal heart sounds, ambient noises, and fetal movements[10]. Additionally, throughout the signal transmission process, the transmission lines and methodologies unavoidably introduce noise interference into the FHR, thus presenting significant challenges in the pursuit of obtaining precise FHR values[11].

In current clinical practice, Doppler ultrasound sensors have become the preferred device for monitoring FHR, also possessing the capability to simultaneously record uterine contractions. This technology is extensively employed in clinical settings to monitor changes in FHR, owing to its low operator skill requirements, high sensitivity to fetal movement signals, and relatively precise data recording[12]. Typically, the monitoring process entails sampling at a rate of four points per second over a continuous period of twenty minutes, allowing for comprehensive observation of the FHR. When deemed necessary, clinicians may suggest extending the monitoring period to forty or sixty minutes to further evaluate the FHR condition[13-16]. A pivotal objective of this monitoring strategy is to precisely identify fetal compromise through meticulous analysis of any decreasing trends in FHR.

In the assessment of fetal compromise, the traditional calculation scheme is currently widely adopted in clinical practice. This scheme primarily relies on two core indicators: the median absolute deviation of the FHR baseline and the variability of the FHR, combined with specific and relatively stable durations, for comprehensive analysis and judgment[17-19]. However, this method inevitably leads to inconsistencies between instrumental automatic assessments and clinical doctors' empirical interpretations[20,21]. The current methods used to detect fetal compromise have relatively low sensitivity, typically fluctuating within the range of 31% to 48%. Additionally, these methods face a high false-positive rate, incorrectly identifying fetuses as being in a compromised state, with this rate ranging between 16% and 21%[22,23]. The inadequacies in accuracy of existing detection methods may result in delays in timely care for infants actually at high risk, as they may not be correctly identified as cases requiring urgent attention[24,25].

At present, the precise determination of fetal heart signal deceleration phases caused by hypoxia has not been adequately and thoroughly investigated. We believe that the main reason for this situation lies in the insufficient accuracy of traditional calculation methods in identifying fetal compromise. These traditional methods often overly rely on fixed algorithms, leading to significant limitations in their identification precision. Relevant studies[26]have further confirmed this, revealing considerable room for improvement in traditional methods within the field of fetal heart signal analysis. As a consequence, we currently lack a dataset that is both highly accurate and representative to support more in-depth research in this area.

Although deep learning models have been applied to fetal heart monitoring tasks, their application effects and popularization are restricted by varying clinical objectives. See the Table 1, The MCNN model analyzes FHR segments from the last 60 minutes before birth to predict post-birth fetal compromise and hypoxia[24]; TG-net aims to classify abnormal groups (umbilical artery pH < 7.20 or 1-minute Apgar score < 7) and normal groups from fetal monitoring data[27]; DeepFHR processes collected FHR signals through wavelet transforms and combines time-domain and frequency-domain features, using Convolutional Neural Networks (CNN)[28] to obtain automatic predictions of fetal acidosis through Bidirectional Long Short-Term Memory (BiLSTM) networks[29]; and there are also approaches that integrate CNN and BiLSTM to process mixed data for the classification of pathological fetuses[30].

**Table 1. Summary of State-of-the-Art Approaches for Fetal Heart Signal Classification from Cardiotocographic Recordings**

| Model Name | Ability for Early Intervention | Acceleration/Deceleration Values | Real-time Variability Analysis | Multitasking Capability |
|---|---|---|---|---|
| MCNN[24] | No | No | No | No |
| TG-net[27] | No | No | No | No |
| DeepFHR[28] | No | No | No | No |
| CNN-BiLSTM[29] | No | No | No | No |
| Mixed-Data | No | No | No | No |
| **FHrCTG(OURS)** | **YES** | **YES** | **YES** | **YES** |

**Table. 1 | General Process of Fetal Heart Rate Monitoring.**

Although existing methods can assist clinicians in intervening during the early stages of labor, thereby preventing permanent damage to the fetus through appropriate means, these methods are typically limited to classifying and assessing the overall condition based on fetal heart signals, with a primary focus on the analysis of static segments[31,32]. Such methods do not address the assessment of real-time FHR variability features, which we believe hold greater clinical significance. These real-time variability features have the potential to be identified at the earliest stages of fetal compromise events, enabling more timely clinical interventions.

In this meticulous and in-depth study, we have carefully constructed a fetal heart signal dataset covering a wide range of time frames. This dataset comprehensively documents FHR data varying from 10 to 60 minutes, with precise annotations of both FHR and its cyclical variations.

To further enhance the accuracy and efficiency of fetal heart monitoring, we have developed a comprehensive fetal heart monitoring model (see Fig. 1c). The core component of this model is an encoding module based on a convolutional neural network, which effectively filters out various interferences in the signal, ensuring precise reconstruction of fetal heart characteristics. Additionally, utilizing an attention mechanism, we have cleverly constructed a dual-decoder module that exhibits excellent performance in feature extraction. Furthermore, during the task discrimination phase, we have innovatively designed an expandable task interface, enabling the model to flexibly learn from and incorporate multiple types of human gynecologists' experiences and judgments on fetal heart signals. This provides more precise and reliable assistance for clinical practice.

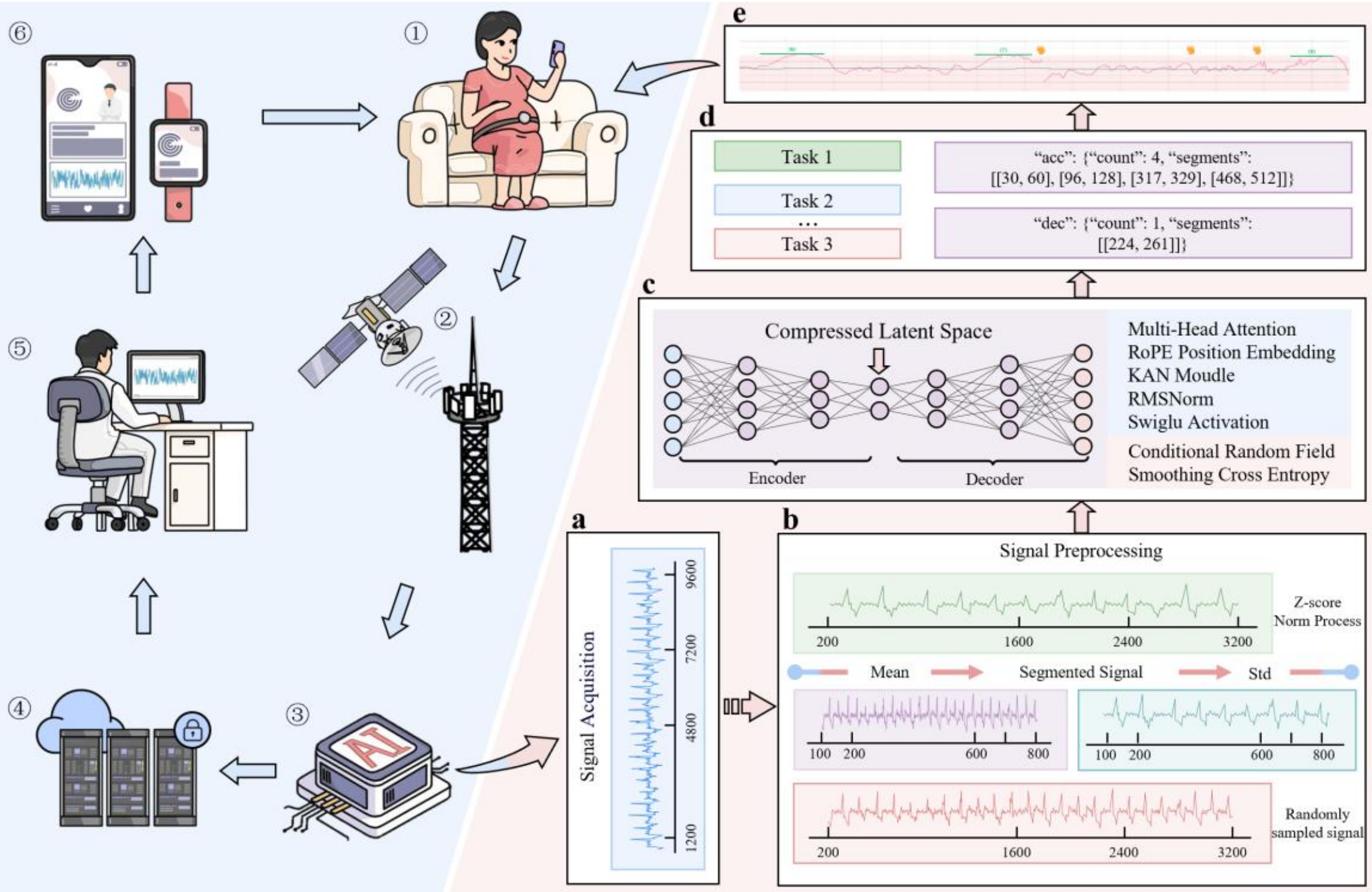

**Fig. 1 | General Process of Fetal Heart Rate Monitoring.** In the left diagram, fetal heart signals are first collected (1) and transmitted to an artificial intelligence model (3) via a signal transmission device (2). The predicted data is then stored (4) and can be accessed by authorized obstetricians, with the results communicated to users through various means (6). we extract statistical information from fixed-length Doppler FHR waveform data **(a)** based on our defined rules, and reconstruct it into waveform data of three different lengths **(b)**. The processed data is then input into the proposed AI model **(c)** for automatic integration and analysis. **d** and **e** illustrate the LABEL format for multi-task outputs and the user interface displayed on the client side.

The model successfully constructed in our study not only offers comprehensive functional coverage and superior performance but also, more significantly, exhibits notable scalability. This attribute underscores the model's extensive applicability and considerable flexibility. Precisely, our model is adept at both stably delivering cloud services in an environment rich in computing resources and being adaptably implemented on portable devices with comparatively limited computing resources, thereby comprehensively meeting the requirements of diverse practical application scenarios.

# Results

## Focus on the Analysis of FHR Monitoring

Our research focuses on the development of an Artificial Intelligence model for FHR monitoring, which aims to maximize fetal safety through its advanced early warning mechanism, promptly detecting and addressing any conditions that may pose a threat to the fetus. It is well-established that fetal hypoxia is a significant factor contributing to miscarriage or future developmental delays, and such hypoxic conditions are often closely associated with attenuations in FHR[3,4,11,13-16]. This study collected and thoroughly analyzed data on the duration of FHR accelerations and decelerations in utero, as detailed in Fig. 2a. The data reveal that the most common duration for heart rate accelerations is 24 seconds, while decelerations typically last for 18 seconds. Notably, both acceleration and deceleration events exhibit normal distribution characteristics, with a noticeable increase in occurrence probability at the 114-second duration mark. Although this phenomenon is intriguing, the clinical significance of this finding is not explored in depth in this study.

To ensure that our model can effectively distinguish between FHR accelerations and decelerations, we have provided ample data support and validation, and employed manual annotation to mark the onset and termination of FHR changes. However, during this process, we also observed that there is no precise criterion for determining the onset and termination of these velocity changes. Even the same expert may exhibit slight variations when annotating the same dataset, and these variations have minimal impact on the overall results (see Fig. 2b). Therefore, base on the practical experience of gynecological experts and their judgments of outcomes, we innovatively propose an Indicator of Overlap (IOL) between predictions and true standards, converting the time series problem into a dichotomous one for validation. Detailed information will be provided in the algorithm development section. Referring to Fig. 2g, the deceleration indicator based on IOL can be observed, while the validation of the acceleration part is demonstrated in Fig. 2h.

## Assessment of FHR Decelerations and Accelerations

In the validation of FHR decelerations, the model constructed in this study demonstrated significant performance advantages. Specifically, when the model predicted a FHR deceleration event, the probability of an actual observed deceleration occurring was as high as 83.88%. Furthermore, the model achieved impressive results in overall prediction accuracy, reaching 88.34%. Its sensitivity index also performed well, at 89.13%, fully demonstrating the superiority of our model in the field of deceleration prediction.

Particularly noteworthy is that the model exhibited higher accuracy in the validation of FHR decelerations (reaching 88.34%, compared to 78.84% for acceleration validation). This significant difference may stem from the prominence of deceleration features in the dataset or be closely related to the specific distribution characteristics of deceleration samples. Additionally, this improvement in accuracy may suggest that the model possesses unique expertise in identifying FHR decelerations. It is worth noting that this expertise may arise from the model's ability to uncover key signal features through deep learning that human doctors have not yet mastered. These findings not only showcase the potential of the model in the field of FHR monitoring but also provide new perspectives and ideas for future AI-based clinical decision support systems.

During the validation process for FHR accelerations, the model similarly exhibited strong predictive capabilities. When the model predicted accelerations, the probability of actual accelerations occurring was approximately 86.38%. However, compared to deceleration validation, the overall accuracy of acceleration validation was slightly lower, meaning that the model correctly predicted approximately 78.84% of the samples in the test set.

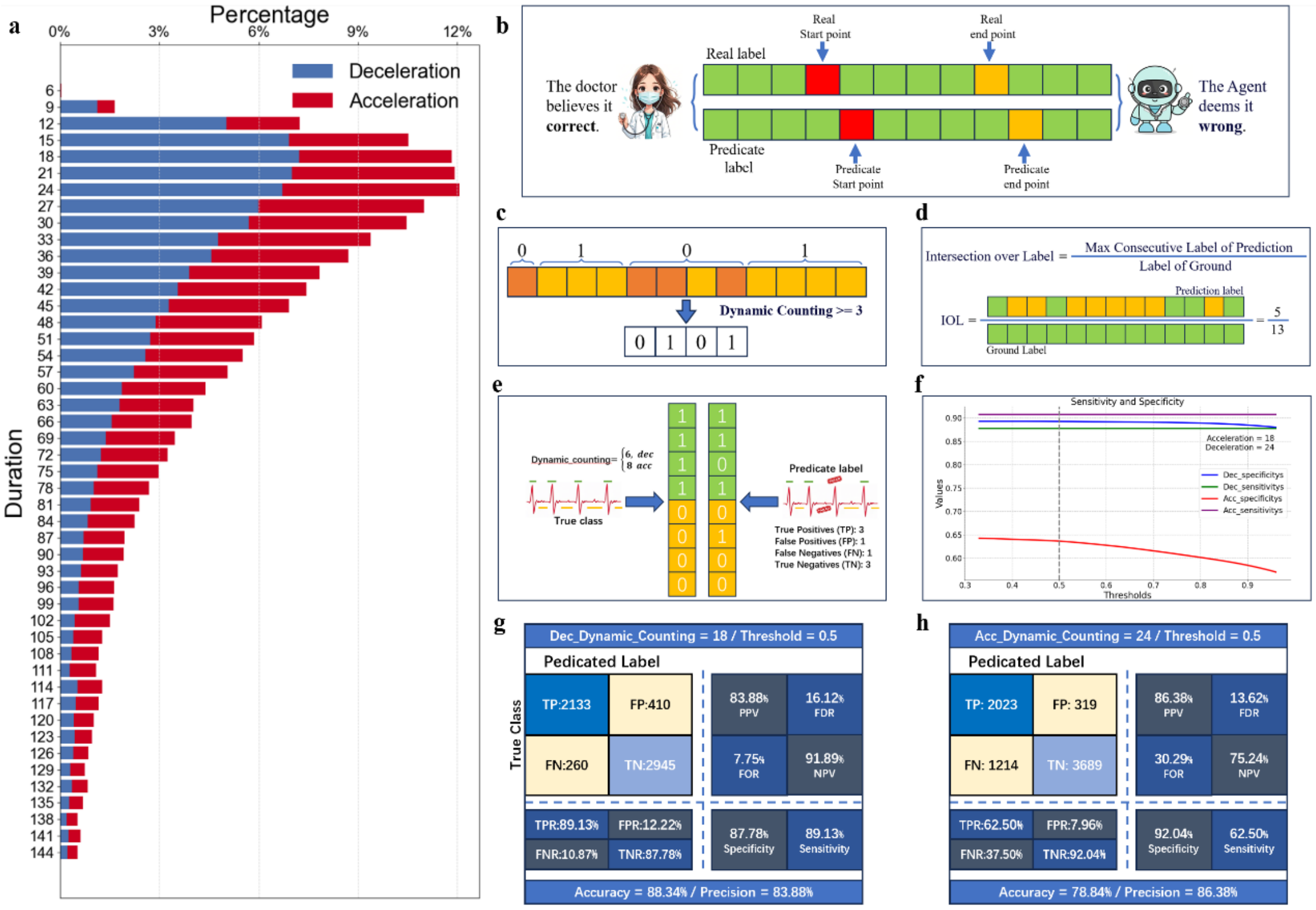

**Fig. 2 | Conclusions Regarding Fetal Heart Rate Acceleration and Deceleration. a** We have statistically analyzed the durations of FHR acceleration (shown in red) and deceleration (shown in blue) in our dataset. The most frequently observed duration for acceleration phases was 24 seconds, while decelerations typically lasted for 18 seconds. **b** show Judgment of results under different objects. **c** and **e** illustrate the process of converting time series data into a binary classification method and subsequently assessing its performance. **d** defines IOL (Intersection over Label) as the ratio of the maximum continuously predicted time segment that overlaps with the annotated label. **f** presents the IOL parameters that yield the optimal sensitivity and specificity across different thresholds. **g** and **h** display the respective metric results on the test set at specific threshold values.

## Assessment of Variability in FHR Monitoring

Periodic variability (PV) and amplitude variability (AV) occupy a pivotal position in FHR monitoring, serving as indispensable indicators for assessing fetal health status, as supported by multiple studies[33-35]. PV, referring to the periodic fluctuations in FHR, provides profound insights into the regularity and stability of fetal cardiac activity. Through meticulous observation of this indicator, professionals can discern potential risks such as fetal hypoxia or compression, enabling prompt and targeted interventions[36].

AV, which represents changes in the amplitude of FHR fluctuations, also contains abundant clinical information. Significant increases or decreases in AV may signify alterations in the intrauterine environment, such as placental dysfunction or intrauterine growth restriction, among other critical conditions. Therefore, precise capture and thorough analysis of AV are crucial for ensuring fetal safety and preventing adverse pregnancy outcomes[37].

From an operational perspective, we designate the values of PV and AV as CPM and BPM, respectively, both of which are ordered along the time axis based on actual fluctuations. These values accurately reflect the fetus's real-time physiological state. In traditional fetal heart signal processing workflows, these data undergo a series of complex algorithmic computations and require manual corrections before being finally displayed on the monitoring interface for pregnant women.

However, this study strives to forge a new path by exploring an innovative model that aims to disrupt the conventional paradigm and replace the cumbersome review tasks typically undertaken by gynecological experts. Our research is not only grounded in traditional numerical definitions but also ingeniously integrates signal feature fusion and reconstruction techniques. This transformative approach simplifies the complex computational problem of PV and AV into a classification task. This innovative shift in thinking allows us to directly and accurately predict the displayed values, ensuring monitoring precision while achieving a significant leap in work efficiency.

## Periodic Variability of Fetal Heart Rate

This study adhered to a progressive principle, rigorously following the classification criteria established by gynecological experts, to conduct an in-depth and detailed analysis of all annotated data. We particularly focused on the proportion of different periodic variability values within the dataset and carefully transformed the results into intuitive graphical representations. In Fig. 3a, we comprehensively present the distribution of these values across the entire annotated dataset. This distribution spans ten distinct categories, which, for ease of identification and reference, are sequentially labeled using the numbers 0-9 and collectively termed "Score Rank" in this paper. Additionally, we emphasized the top seven values with the highest proportions, aiming to assist readers in swiftly grasping the core characteristics of the data.

Our deep learning model demonstrated satisfactory predictive performance on the test set. Fig. 3b provides a detailed illustration of the prediction results on the test set, where both the x-axis and y-axis of the graph represent the counted CPM values. Upon close examination of the graph, it becomes evident that the degree of dispersion in the prediction results adjusts accordingly with changes in the data volume. In the corresponding violin plot, we can clearly observe significant differences in the data distribution density across different prediction categories, with a tendency for the data to cluster towards the center. However, it is noteworthy that this performance may have been influenced by an imbalance in the distribution of the training data, a crucial factor that must be considered in future model optimization efforts.

Although the variation periods of fetal heart signals can be clearly defined, calculated using statistical methods, and expressed as specific numerical values, in actual clinical practice, we often rely on the widely applied Fischer evaluation method beyond these basic parameters. Specifically, we primarily consider the "binned" intervals of fetal heart periods based on the "binning" formula provided by the Fischer scoring system[38]:

$$CPM\ Bucketing\ Score\ = \begin{cases} 0, cpm < 2 \\ 1, cpm \in [2,5] \\ 2, cpm > 5 \end{cases}$$

Fig. 3c presents the detailed results of our "binning" process for the model-predicted variation periods, combined with Fischer scores. Following the Fischer criteria, we subdivided the variation period values into three intervals. In this figure, we maintain the same coordinate system as in Fig. 3b, where the x-axis represents different types of model predictions, and the off-diagonal positions reflect prediction errors. The size of the spheres in the figure is determined by the ratio of the current number of predictions to the total predictions.

From the graphical representation, it is evident that areas with higher data density are concentrated near the boundaries. This phenomenon may be influenced by the manual scoring strategies employed by gynecological experts during data calibration. Experts may tend to adopt a more neutral scoring approach for certain data, leading to a certain bias in the data. This is an important issue we identified during the experimental process, but it will not be discussed in this paper.

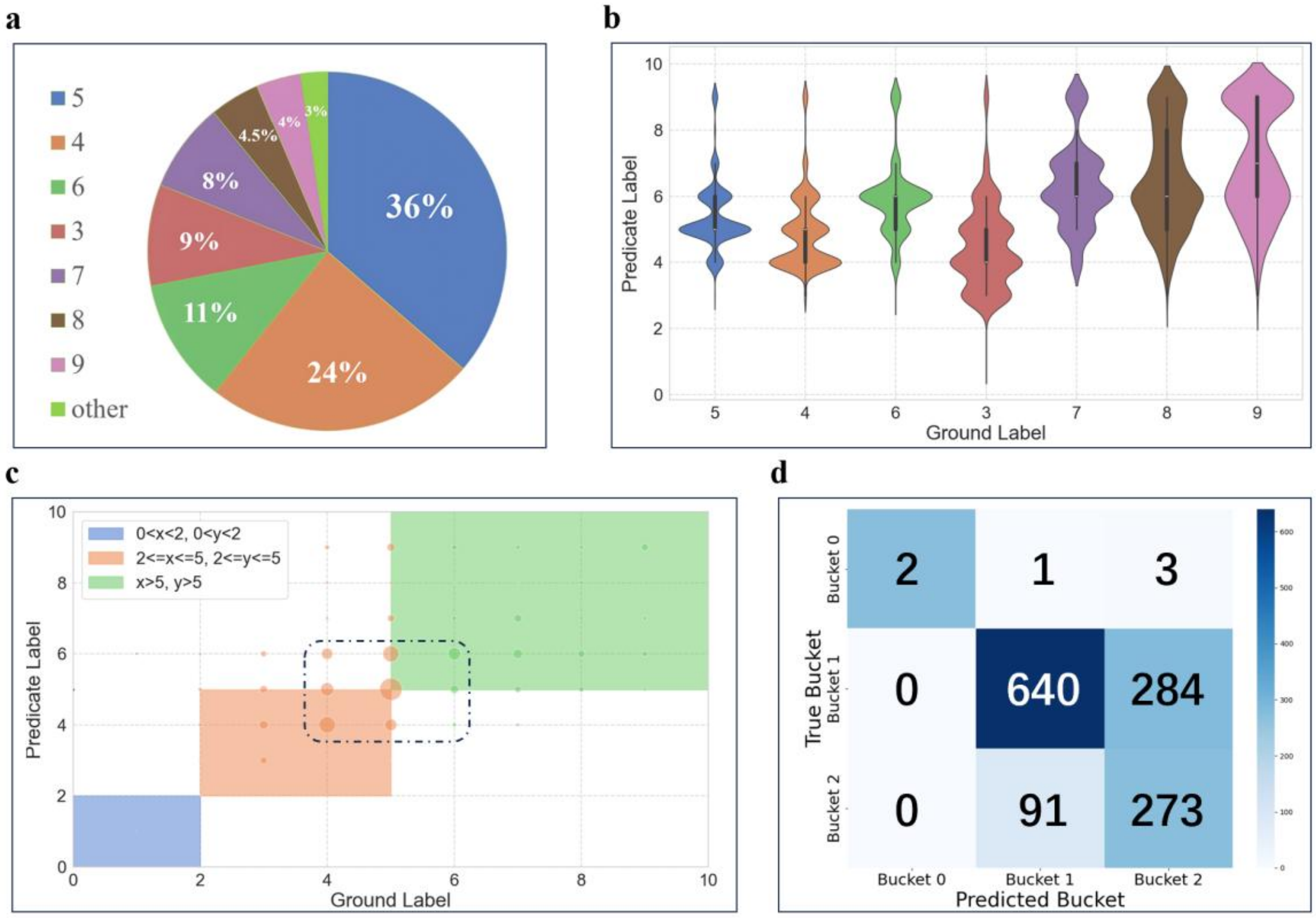

**Fig. 3 | Conclusions Regarding Periodic Variability. a** meticulously displays the distribution of various cycle variability values within the annotated dataset, illustrating the proportion of each. Building upon the framework of baseline prediction scores, **b** profoundly unveils the distribution pattern of model predictions. Furthermore, **c** precisely categorizes the predicted values into various bucketing intervals based on the Fischer score. Lastly, **d** conducts a comprehensive and in-depth analysis of prediction accuracy utilizing a confusion matrix grounded in the Fischer scoring method.

Upon deep analysis, it was revealed that despite the significantly larger data volume of CPM4 compared to CPM6, the model exhibited an overall tendency to overestimate when predicting CPM5 calibrations, which is inconsistent with the proportional distribution of the datasets. Therefore, we postulate that this issue is not solely related to the datasets themselves but may also involve deeper knowledge in gynecological assessments that we have not yet fully comprehended or mastered. This finding, to a certain extent, demonstrates that our model can effectively capture the true state and has successfully learned from the experiences and knowledge of gynecological experts during the later calibration process.

Additionally, Fig. 3d presents a confusion matrix constructed based on the Fischer scoring method, providing further detailed insights into the model's performance. For a more comprehensive data analysis, please refer to the tabular data in Fig. 4c.

## Amplitude Variability of Fetal Heart Rate

In Fig. 4a, we meticulously illustrate the overall distribution of the training dataset and conduct a detailed comparison with our model's predictions on the test dataset. In the right-hand section of Fig. 4a, we specifically highlight the distribution of amplitude variability across various bucketing intervals. The figure clearly demonstrates that the data points predominantly concentrate within the [BPM11, BPM22] interval, aligning with a standard normal distribution pattern. Notably, a dense clustering of data points is also observed at the BPM9 marker. We postulate that this significant increase is closely associated with the bucketing mechanism employed by the Fischer method for scoring amplitude variability. The relevant formula is presented below:

$$BPM\ Bucketing\ Score = \begin{cases} 0, bpm < 5 \\ 1, bpm \in [5,9]\ or\ bpm > 30 \\ 2, bpm \in [10,30] \end{cases}$$

In the left diagonal section of Fig. 4a, we meticulously present the model's predictions of amplitude variability values. The gray spheres in the figure represent the offset of the current predicted values, which profoundly reflect the error and bias between the predicted and actual values in the test set. Notably, within the specially annotated box, the overall offset trend of the predicted values significantly converges towards the center of the original distribution. This finding is highly consistent with the overall numerical distribution characteristics of amplitude variability, providing valuable insights for us to further understand model behavior and optimize prediction strategies.

To comprehensively evaluate the model's prediction performance, we conducted detailed statistical analyses of the prediction results across different BPM values. Fig. 4b displays a selection of representative samples of predicted amplitude variability values. The results indicate that the model exhibits the highest recall rate at the BPM9 annotation, demonstrating its effective ability to identify positive samples under this annotation. Meanwhile, the highest precision rate is observed at the BPM24 annotation, meaning that the proportion of true positive samples among the model's predictions is the highest under this condition, reflecting the model's excellent performance in this scenario. Detailed assessment data using the Fischer binning method are presented in the table in Fig. 4d for readers' reference.

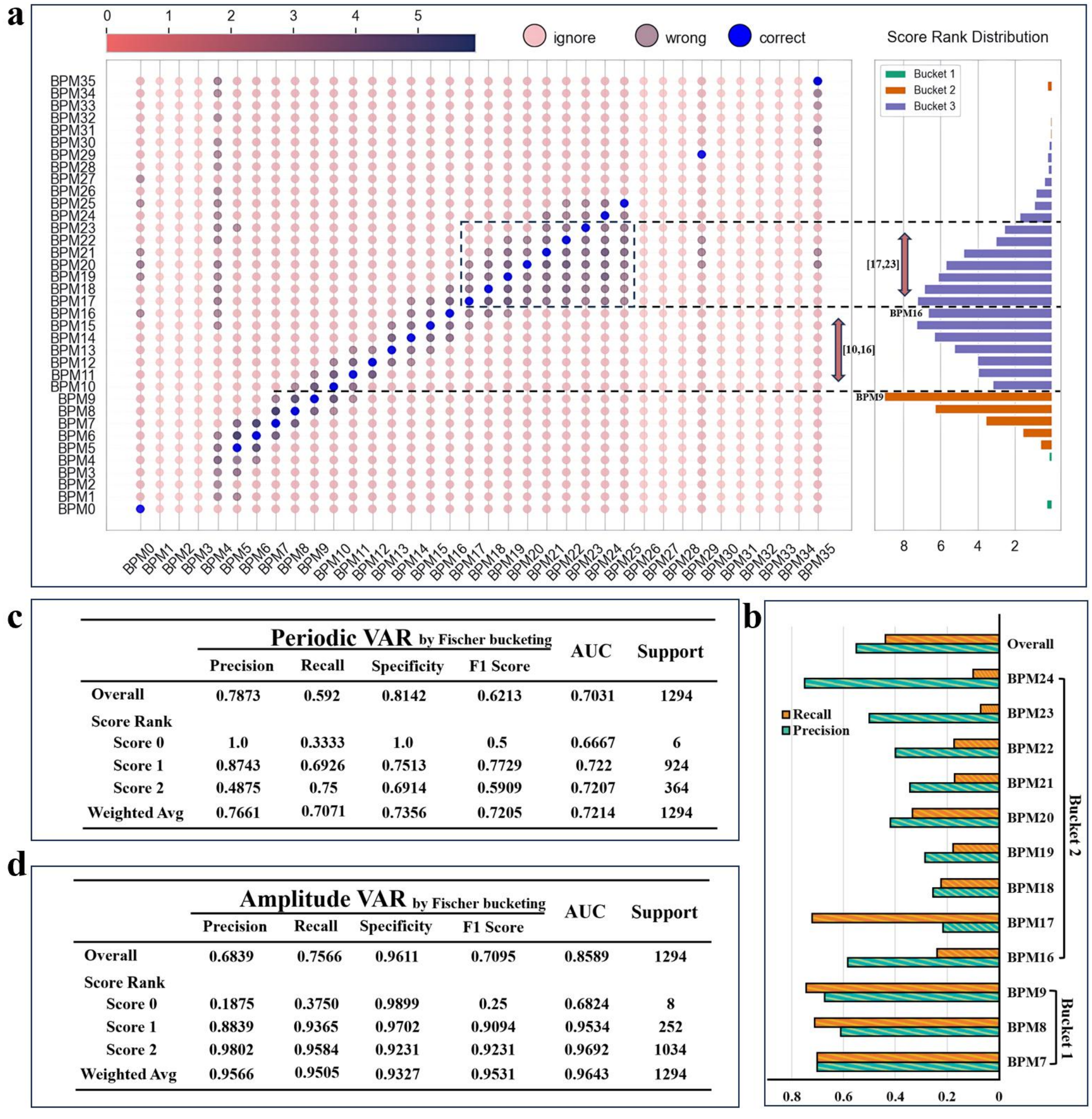

| | Periodic VAR by Fischer bucketing | | | | AUC | Support |
|---|---|---|---|---|---|---|
| | Precision | Recall | Specificity | F1 Score | | |
| Overall | 0.7873 | 0.592 | 0.8142 | 0.6213 | 0.7031 | 1294 |
| Score Rank | | | | | | |
| Score 0 | 1.0 | 0.3333 | 1.0 | 0.5 | 0.6667 | 6 |
| Score 1 | 0.8743 | 0.6926 | 0.7513 | 0.7729 | 0.722 | 924 |
| Score 2 | 0.4875 | 0.75 | 0.6914 | 0.5909 | 0.7207 | 364 |
| Weighted Avg | 0.7661 | 0.7071 | 0.7356 | 0.7205 | 0.7214 | 1294 |

| | Amplitude VAR by Fischer bucketing | | | | AUC | Support |
|---|---|---|---|---|---|---|
| | Precision | Recall | Specificity | F1 Score | | |
| Overall | 0.6839 | 0.7566 | 0.9611 | 0.7095 | 0.8589 | 1294 |
| Score Rank | | | | | | |
| Score 0 | 0.1875 | 0.3750 | 0.9899 | 0.25 | 0.6824 | 8 |
| Score 1 | 0.8839 | 0.9365 | 0.9702 | 0.9094 | 0.9534 | 252 |
| Score 2 | 0.9802 | 0.9584 | 0.9231 | 0.9231 | 0.9692 | 1034 |
| Weighted Avg | 0.9566 | 0.9505 | 0.9327 | 0.9531 | 0.9643 | 1294 |

**Fig. 4 | Conclusions Regarding Periodic Variability.** The right **(a)** displays the distribution of BPM across the entire dataset, with color coding based on Fischer binning method. The left presents the prediction results of our model on the test set, where gray indicates incorrect predictions. **b** illustrates the precision and recall of the model's predictions of amplitude variability across major BPM categories. **c** and **d** display the evaluation metrics for predicted period variability and amplitude variability, respectively, after applying the Fischer method for BUCKETING.

Overall, our artificial intelligence system for FHR monitoring developed by our research team is dedicated to maximizing fetal safety. The core mission of this system is to establish an early warning mechanism that can promptly detect and respond to any conditions that may pose a threat to the fetus. Additionally, we innovatively propose the Indicator of Overlap (IOL) between AI prediction results and true standards as a crucial metric for evaluating model performance.

During the validation phase, the AI model demonstrated impressive performance in predicting both FHR decelerations and accelerations. Its accuracy and precision reached high levels, particularly in identifying FHR decelerations, where the model exhibited exceptional performance. This achievement is of utmost importance in reducing misdiagnosis rates and ensuring fetal safety.

Furthermore, this study delved into the pivotal roles of periodic variability and amplitude variability in FHR monitoring and fully demonstrated our model's predictive capabilities for these two core indicators. Accurate capture and analysis of periodic variability and amplitude variability are crucial for comprehensively assessing fetal health status and effectively preventing adverse pregnancy outcomes. In predicting these variability indicators, our model exhibited strong learning ability and accurate grasp of true states. These achievements, based on deep learning, undoubtedly provide valuable learning opportunities and reflective space for humans.Indeed, artificial intelligence can enhance FHR monitoring, paving the way for safer and more effective prenatal care.

# Discussion

## Core Findings and Model Innovation

In this study, we developed an end-to-end deep learning model, FHrCTG, tailored for Doppler fetal heart signal analysis, aiming to deliver highly accurate and reliable diagnostic outcomes. The model streamlines the fetal heart signal processing pipeline into three stages: signal acquisition, transmission, and analytical output (Fig. 1b). By integrating pre-training technology and a novel fetal heart signal decomposition-reconstruction method, FHrCTG effectively extracts features under interference conditions, enhancing feature density and focusing on core analysis tasks. This innovation addresses the critical challenge of ensuring signal fidelity in complex environments, where transmitted signals must accurately reflect fetal status to maintain usability.

## Performance Analysis: Precision, Accuracy, and Clinical Trade-offs

Our evaluation revealed nuanced performance differences between accelerated and decelerated predictions. During accelerated validation, the model achieved high precision (86.38%) but lower accuracy (78.84%), suggesting a tendency to misclassify non-accelerated samples as accelerated. This discrepancy may stem from the relatively lenient criteria for fetal heart acceleration in clinical diagnostics, leading to annotation flexibility. Conversely, deceleration predictions demonstrated superior precision and accuracy, highlighting the model’s robustness in identifying critical fetal distress indicators.
Further analysis using sensitivity and specificity metrics showed that FHrCTG achieves 89.13% sensitivity and 87.78% specificity in deceleration detection, significantly outperforming its acceleration prediction performance (62.50% sensitivity, 92.04% specificity). This aligns with clinical priorities: minimizing missed diagnoses of decelerations (high sensitivity) is critical, even at the cost of increased false positives. Notably, the model's specificity remains strong in both modes, indicating its ability to exclude false positives, particularly in acceleration predictions (Fig. 2g&h).

## Clinical Implications and Scalability

The FHrCTG model demonstrates exceptional scalability across hardware and software platforms. It can operate on large servers for professional diagnostics or adapt to portable devices in remote or emergency settings, independent of power or wireless networks. This flexibility supports prenatal care in underserved regions. Additionally, diversified task interfaces enable adaptation to various fetal heart analysis tasks, providing technical support for broader research applications.

Clinically, the model serves as a quality benchmark for gynecologists and a tool for independent practice. In resource-limited areas, it assists in timely fetal biometric assessments, including FHR changes, periodic variability, and amplitude variability—key indicators for diagnosing fetal growth conditions. Furthermore, our model introduces standardized evaluation criteria for acceleration/deceleration judgments, addressing gaps in existing fetal monitoring frameworks.

## Model Generalization and Scoring System Performance

Compared to clinician-reported assessments, FHrCTG reduces discrepancies with clinical reality, demonstrating superior generalizability. In the Fischer scoring system—a cornerstone of clinical practice—the model excels in amplitude variation analysis, achieving high precision and recall in weighted average indicators (Fig. 3c&d). This success stems from mitigating data imbalance across score ranks during training, enhancing real-world reliability.

## Limitations and Future Improvements

Despite its strengths, FHrCTG faces challenges in periodic variation recognition. Evaluation metrics for cyclic parameter monitoring (CPM) annotations remain suboptimal, with F1 scores below 0.8 in all scenarios. This highlights the need for improved handling of periodic patterns, potentially through refined data annotation protocols or advanced temporal modeling techniques.

## Conclusion and Broader Impact

The FHrCTG model represents a pioneering application of AI in fetal heart monitoring, establishing evaluation standards for critical detection indicators. Its intuitive visualization interface (Fig. 1e) enables direct use by pregnant women via remote support, democratizing access to autonomous fetal monitoring. By bridging technological innovation and clinical practice, this work advances prenatal healthcare quality and accessibility, offering a transformative tool for both medical professionals and patients. Future efforts will focus on optimizing periodic variation analysis and validating the model in diverse clinical settings.

## Study on the Analysis of FHR Monitoring

After a thorough comparison of the model's precision and accuracy, we observed a noteworthy phenomenon. During the accelerated validation phase, while the model demonstrated a high precision of 86.38%, its accuracy remained at 78.84%. This discrepancy may reveal a specific tendency of the model in handling accelerated predictions: it can identify most true accelerated samples with considerable precision, but there may also be some misjudgments, wherein a portion of non-accelerated samples are incorrectly classified as accelerated. We speculate that this phenomenon might be related to the relatively lenient criteria for fetal heart acceleration in obstetric diagnostics, leading to some flexibility and broadness in data annotation. In contrast, during the decelerated validation phase, the model exhibited excellent performance in both precision and accuracy, fully demonstrating its superior performance in this domain.

Further assessing the model from the dimensions of sensitivity and specificity, our model showed a sensitivity of 62.50% and a high specificity of 92.04% in accelerated predictions; whereas in the field of decelerated predictions, these two key indicators reached 89.13% and 87.78%, respectively. This set of data clearly indicates that the model exhibits higher sensitivity in capturing and predicting decelerated events (89.13% compared to 62.50%, representing a significant improvement), while its performance is slightly weaker in predicting accelerated events. This finding aligns closely with our expectations for the model, namely that in the task of fetal heart deceleration identification, the model must possess an outstanding true positive rate to minimize the likelihood of missed diagnoses, even at the potential cost of increasing the risk of misdiagnosis. It is also noteworthy that although the model demonstrates good specificity in both prediction modes, its performance in accelerated predictions is slightly superior to that in decelerated predictions, indicating that the model is more proficient at excluding false positive results when predicting accelerated events.

# Methods

This study encompasses five core concepts that are both innovative and possess practical application potential.

These include data source selection, data preprocessing procedures, model design schemes, model pre-training methods, and the novel IOL algorithm developed in this study. Next, we will delve into each of these concepts in detail.

## Dataset

The dataset utilized in this study is derived from information gathered and compiled by portable Doppler FHR monitoring devices supplied between 2020 and 2024. All data have undergone meticulous manual verification to ensure accuracy and have been extensively employed for model training, validation, and internal testing. Throughout the model training phase, we screened a total of 558,412 FHR records, with 7,266 of these containing detailed recordings of accelerations and decelerations. To guarantee the precision of experimental conclusions, we meticulously selected 689 annotated records from the entire dataset to serve solely as a test set, ensuring no overlap between the test and training sets.

The process of data annotation is of utmost importance. We enlisted 17 professionals, comprising 3 experienced gynecologists and 14 medical interns, for this task. Each FHR record underwent thorough review and re-annotation by two medical interns, and any discrepant annotations were reviewed by the gynecologist team to finalize the annotation results. This rigorous and detailed annotation protocol, tailored specifically for this research endeavor, aims to bolster model reliability and ensure the objectivity and accuracy of evaluation outcomes.

The annotation endeavor was aided by a specialized tool developed through secondary development based on the open-source Label Studio. Each annotation precisely documents the start time, end time, and duration of FHR accelerations and decelerations. Furthermore, we comprehensively recorded the FHR baseline, along with the maximum amplitude variation and maximum period variation relative to this baseline.

It is noteworthy that, although our collected data also encompassed uterine contraction signals, our extensive testing and detailed analysis of existing records revealed considerable controversy surrounding the collection methods of uterine contraction signals and their reliability in clinical decision-making[39]. Consequently, in the model design of this study, we opted to exclude this feature temporarily.

The FHR signals we gathered encompass a diverse range of sources and users, aiming to more accurately reflect fetal conditions and data characteristics across various clinical and real-world medical scenarios. By leveraging these representative datasets for model training and analysis, we can more accurately assess model performance and draw conclusions with practical implications. Simultaneously, this approach markedly enhances the credibility of our study.

## Data pre-processing

Based on profound insights into the field of deep learning, this study is dedicated to constructing an end-to-end deep learning model for efficient processing of fetal heart signals. The core advantage of this model lies in its ability to directly parse and reconstruct raw acquired signals, demonstrating not only robust anti-interference capabilities but also maximizing the retention of crucial information within the signals. To achieve this goal, we innovatively propose a novel sequence scaling method. This method delicately aligns input features of varying lengths, thereby greatly simplifying the complexity of feature alignment. Simultaneously, as a significant technological innovation in the field of feature fusion, this method markedly enhances the overall performance of the model, providing a more powerful and precise tool for the processing and analysis of fetal heart signals.

At the implementation level, we focus on fetal heart signals as the core input feature and employ a series of meticulous preprocessing steps aimed at significantly boosting the model's efficacy. Inspired by the successful application of multimodal data fusion in various domains[40], our model adopts four carefully processed features as inputs. Specifically, this study initially conducts precise sampling of fetal heart time signals (see Fig. 1a), followed by segmentation through multiple windows. To enhance signal reconstruction efficacy, we calculate the corresponding mean and variance within each window, aiming to utilize these statistics to intuitively represent the central tendency and dispersion of signals in each window (see Fig. 1b). Furthermore, we introduce the z-score to accurately depict the relative position of the current value within the mean of its respective window. It is noteworthy that the continuous fetal heart signal sampling strategy and the novel calculation method for multi-window statistics employed in this study are pioneering in the field of fetal heart signal processing, constituting an innovation in data processing and fusion. The following section will elaborate on the preparatory workflows for each branch of input.

We employ a windowing strategy to process the raw FHR sequence data, which segments the entire sequence into multiple parts using fixed-size windows. Specifically, let the original fetal heart signal sequence be denoted as $X = \{x_1, x_2, \ldots, x_n\}$. In this study, we select a window size of 12 units, i.e., $w = 12$,to partition $X$. This step can be formally represented as dividing $X$ into multiple subsequences $S_i$, where each subsequence contains $w$ consecutive elements, i.e.,$S_i = \{x_{i,1}, x_{i,2}, \ldots, x_{i,w}\}$,for $i = 1,2, \ldots, m$,where $m$ is the number of subsequences after partitioning.

Next, we calculate the mean $\mu_i$ of the sequence within each window, which also reflects the average intensity of the fetal heart signal within that window. The formula for calculating the mean $\mu_i$ is as follows:

$$\mu_i = \frac{1}{w} \sum_{j=(i-1)w}^{iw} x_j$$

Through this computation, we obtain a new integer sequence $M = \{\mu_1, \mu_2, \ldots, \mu_m\}$ , where each element represents the mean of the fetal heart signal within the corresponding window. Similarly, we also generate a variance value for each window to capture the variability of the fetal heart signal within that window. The formula for calculating the variance $\sigma_i^2$ is as follows:

$$\sigma_i = \frac{1}{w}\sum_{j=(i-1)w}^{iw}(x_j - \mu_i)^2$$

This will generate another new sequence $V = \{\sigma_1, \sigma_2, \dots, \sigma_m\}$, where each element represents the variance of the fetal heart signal within the corresponding window. The core of this strategy lies in reducing the length of the original sequence through a windowing approach, while preserving the key features of the fetal heart signal. The mean sequence $M$ and variance sequence $V$, as statistical features, not only reveal the signal's intensity but also reflect its variability. These features provide rich and representative informational inputs for subsequent models, aiding in more accurately capturing the underlying patterns and variability within the fetal heart signal.

Furthermore, to substantially enhance the model's accuracy and generalization performance in recognizing fetal heart signals, we have adopted the standard score (i.e., Z-score) as a crucial input feature. As a statistical measure, the standard score can effectively eliminate individual differences and noise interference introduced by measurement devices from the raw signal through a standardization process, and precisely calibrate the relative position of the signal. This processing method is vital for ensuring that the model maintains high adaptability in the face of diverse data variations. In calculating the standard score, we perform the following operations for each analysis window: firstly, the mean value within the window is subtracted from the raw signal to eliminate differences in the signal's average level; secondly, the result is divided by the standard deviation within the window to eliminate differences in the signal's fluctuation amplitude:

$$z_j = \frac{x_i - \mu_i}{\sigma_i}$$

Through this refined operation, we ensure that all input signals are compared on a unified scale, thereby significantly enhancing the model's adaptability to diverse data variations. Especially when the signal acquisition equipment is subject to strong interference or data breakpoints occur, the introduction of standard scores effectively ensures smooth data transitions and processing, maintaining the accuracy and continuity of model analysis.

The data processed with standard scores maintain the same signal length as the original signals. To further shorten the length of the input feature signals while preserving the characteristics of the original signals, we employ a small-window random sampling method. Specifically, we set the size of the random small window to 3 and the sampling value to 1, meaning that one data point is randomly retained as a sampling signal for every three data points. This random sampling approach not only preserves the original signal characteristics but also allows for modifications to the same original signal during each training session, thereby increasing the model's generalization ability for feature signals. We apply this sampling method to both the original signals and the z-score values obtained after standard score processing.

In terms of feature fusion strategy, we carefully designed the ratio of the sizes of the large and small windows to the length of the original signal, setting it at 1:4, which aligns perfectly with the initial 4Hz sampling frequency. This design enables the model to utilize the reshape technique for folding the input signals and subsequent fusion operations. Specifically, the size of the first large window is set to 12, and after sampling, what is captured are essentially the mean and variance of the Doppler signal data over 3 seconds. During the random sampling process, we use a window size of 3, and after the model's internal reshape dimensional adjustment, the concat technique can be utilized for efficient fusion of all input signal features.

This strategy not only achieves effective compression of the feature space but also ensures that the model is exposed to a rich variety of signal samples during the training phase due to the introduction of random sampling. Since each random sampling generates new data combinations, this greatly enhances the model's generalization ability and prediction accuracy for unknown data. Through this ingenious feature fusion technique, our model demonstrates outstanding performance and high accuracy in processing variable and complex fetal heart signals.

## FHrCTG (Fetal Heart Rate Cardiotocography) Architecture

This study innovatively constructs an extensible deep learning FHR monitoring model named FHrCTG (see Fig. 5a). This model efficiently receives and processes Doppler ultrasound signals, and performs diversified FHR monitoring tasks based on reconstructed signal features. We have implemented this model using the PyTorch platform, endowing it with robust functionality and flexibility. Currently, the FHrCTG model has developed three output branches of two types, which can accurately predict FHR acceleration and deceleration periods, as well as periodic and amplitude variations, thereby providing a more comprehensive and in-depth analysis tool for FHR monitoring.

The core innovation of our research lies in its unique Encoder-DualDecoder design. This architecture ingeniously employs a convolutional neural network as the backbone of the encoder, seamlessly integrating it with a decoder that utilizes an attention mechanism[41] .We have implemented residual connections within all modules, achieving efficient concatenation and deep fusion of features[42]. We considered all the extracted features as independent scalar features and employed both a conditional random field (CRF) and softmax for in-depth analysis and determination of the feature results.

Structurally, it comprises four primary components: the ConvBlock for initial feature extraction and signal reconstruction, the DecoderBlock for feature matching, the PostDecoderBlock for secondary filtering, and the output layer for result prediction. All the detailed operations are depicted as follows:

## ConvBLock

The originally acquired signals are often susceptible to interference from various noise sources. These noises mainly originate from the movements of the fetus and the pregnant woman, as well as changes in the acquisition site, which inevitably contaminate the signals obtained through Doppler ultrasound technology[43]. To address this issue, we have designed a convolution-based MergeEncoder (see Fig. 5b), whose core function is to effectively denoise and reconstruct the input data. Through a series of meticulous merging and separating operations, this encoder can reshape and highlight the key features in the original signals. It is worth mentioning that this study is the first to integrate the denoising and reconstruction of one-dimensional discrete time series signals within a single deep learning module, representing a novel innovation in the relevant research field. Furthermore, we have chosen convolutional layers as the backbone network for feature reconstruction, a choice that significantly enhances the overall model's operational efficiency. We have named this core component the "Multi-Windows Sampling and Remodeler" to emphasize its unique functions in multi-window sampling and signal remodeling.

To effectively suppress redundant information in the reconstructed feature space and channels, this study ingeniously integrates plug-and-play SRU (Separated Reconstruction Unit) and CRU (Channel Refinement Unit) modules[44]. Specifically, the SRU targetedly weakens redundancy in the spatial dimension by implementing a separated reconstruction strategy, while the CRU precisely reduces inter-channel redundancy through a multi-step approach of segmentation-transformation-fusion. This clever design not only significantly enhances model performance by eliminating redundant features but also demonstrates notable advantages in substantially reducing algorithm complexity and computational costs.

In the processing of fetal heart signals, we advocate treating them as discrete signals rather than continuous signals, as this more accurately represents the amplitude of the fetal heart in the current time series. After converting these signals into feature vectors using an Embedding layer, we utilize a convolution-based encoder layer and fuse feature values of the fetal heart signals across different statistical measures by constructing cross-channel expressions and connections at various stages (Spatial and Channel Reconstruction for Feature Redundancy). This innovative approach to reshaping the original signals constitutes another major innovation of this study.

We consider the incorporation of absolute positional information to be a crucial step. The raw fetal heart signals lack an understanding of temporal features, and we hope that the reconstructed fetal heart signal features can assign a unique position vector to each location. These vectors are fixed and independent of the content of the input sequence. Absolute positional encoding generates a unique encoding vector for each input position, which contains specific information about the position within the sequence, thereby accurately representing the state of each position. Our formula for absolute positional encoding is as follows:

$$Y = W(x_t + pos_t) + b$$

$$pos_{t,2i} = sin(\frac{t}{L^{\frac{2i}{d}}})$$

$$pos_{t,2i+1} = cos(\frac{t}{L^{\frac{2i}{d}}})$$

$x_t$ represents the input reconstructed feature signal, where $W \in R^{d \times d}$ and $b \in R^{d \times d}$ are learnable parametric matrices. $pos_t$ denotes the absolute positional vector, $i$ is the positional index, while $d$ is the vector dimension. We employ a combination of sine and cosine functions to create encodings for each position, with wavelengths arranged in a geometric progression, ensuring that the encoding for each position is unique. Furthermore, the periodic nature of sine and cosine functions guarantees that the encoding relationships between different positions are continuous and smooth.

## Dual-Decoding

To further enhance the depth and precision of feature extraction, we innovatively adopt a Dual-Decoding architecture and implement a deep supervision strategy, enabling a more comprehensive capture of the encoder's reconstruction of FHR signals. Specifically, the core component of the first decoder is a multi-head self-attention mechanism (see Fig. 5c), which is inspired by previous research[41]. We anticipate that the excellent performance of this self-attention model will enable precise capture of feature information both locally and globally. Furthermore, to further strengthen the feature extraction capability of this mechanism, we have deeply optimized and expanded it by introducing a Feedforward layer with the Swiglu activation function[45]. Additionally, considering the importance of positional information in FHR signal analysis, we deliberately incorporate Relative Positional Encoding (RoPE) at the decoder level. This measure aims to help the model more sensitively perceive and capture key position-related information, thereby enhancing the accuracy of information acquisition[46].

In response to the diversity of FHR observation signals in terms of scale, we cleverly introduce an innovative multi-layer convolutional module at the backend of feature extraction, named "PostDecoderBlock" (see Fig. 5d). Through a meticulously designed structure, this module interleaves one-dimensional convolutional layers with fully connected layers, forming an efficient and adaptive signal filter. The core function of this filter is to meticulously capture and extract signal features across different channels, and then utilize these features to reconstruct time series information. With this design strategy, we expect the model to exhibit higher flexibility and adaptability when dealing with complex and variable FHR signal scales, thereby significantly improving the accuracy of data analysis and optimizing work efficiency.

The ingenious construction of the dual decoders allows the model to process richer feature inputs, which has been practically verified[47]. This innovative design not only enhances the model's flexibility but also lays a solid foundation for its adaptability in broader application scenarios.

## Multi-task prediction

At the model's output stage, our objective is to construct a versatile model capable of broadly adapting to various tasks in fetal heart monitoring. To this end, we have carefully selected the KAN module[48] and the RepConv module (see Fig. 5e) to perform refined modeling of time series features and classification features, respectively. The uniqueness of the KAN module lies in its adoption of spline-parameterized univariate functions, an innovative approach that replaces traditional linear weights, enabling the model to dynamically capture activation patterns. This design strategy not only significantly improves the accuracy of prediction results but also substantially reduces the number of parameters required for learning, thereby achieving efficient operation and practicality of the model. Specifically, the KAN module can map high-dimensional feature vectors to a specific low-dimensional space to accurately represent the onset, termination, and duration of different stages of fetal heart signals (see Fig. 1d). To further enhance the model's ability to distinguish time intervals, we have specifically chosen Conditional Random Fields (CRF) as the decoder in this module, rather than traditional classification methods. Through this improvement, each position's feature vector can obtain a unique identifier for the duration stage of the fetal heart signal.

Meanwhile, the RepConv module processes feature vectors more finely through the stacking of multiple convolutional layers. This structure ensures efficient compression and optimization of the signal while preserving its features intact. The flattened features are then further passed to two independent KAN layers for deep feature processing and extraction, maximizing the mining and utilization of the intrinsic value of the features.

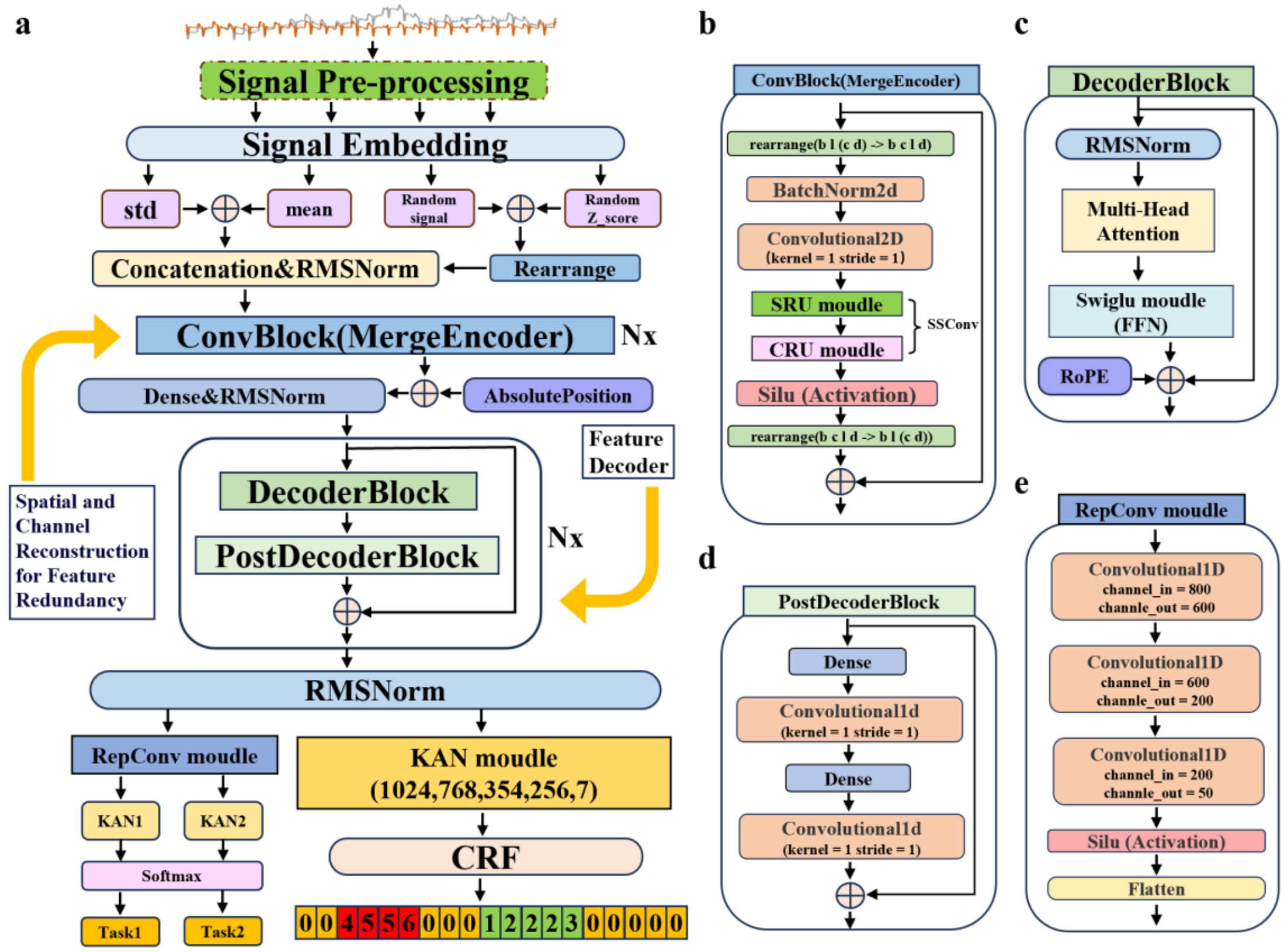

**Fig. 5 | Illustration of Our Proposed Artificial Intelligence Model, FHrCTG.** The left panel depicts the overarching architecture of our model, distinguished by feature fusion and reconstruction, alongside its capability to manage two distinct types of tasks. The right panel elaborates on the diverse components, as expounded in the FHrCTG ARCHITECTURE section of this paper.

In the feature classification task, we employed cross-entropy as the loss function to train our designed model. By integrating the Softmax output function with logistic regression, we implemented an efficient supervised learning mechanism aimed at maximizing the prediction accuracy of label probabilities. This approach demonstrated marked advantages in handling sequence problems involving multiple categories, allowing for a more precise evaluation of the model's performance in sequence analysis tasks. As the cross-entropy loss was continuously minimized, our model gradually learned the discrete representations of fetal heart signals and achieved significant improvements in identifying critical states such as fetal distress. This achievement provides solid support for the clinical application of fetal heart monitoring, with the potential to enhance diagnostic efficiency and accuracy.

## Model pretrain

The feature extraction of the overall model is derived from the Doppler ultrasound fetal heart signal collector, an advanced device capable of precisely capturing critical fetal heart signals. We have meticulously constructed the model to achieve efficient extraction, accurate reconstruction, and comprehensive integration of deep semantic features within these signals. To attain this goal, we adopted an end-to-end training paradigm, enabling the model to directly receive input data for prediction and output the final results, thereby eliminating cumbersome intermediate processing steps.

Given the scarcity of annotated data, we introduced a pre-training strategy to enhance the model's robustness and scalability[49]. During the pre-training phase, we initially truncated random raw data to ensure that the obtained sequences matched the length of the data to be predicted. Subsequently, we randomly masked 4 to 12 data points, equivalent to masking 1 to 3 seconds of the acquired signal, while ensuring that the masked data points did not exceed 15% of a single data sequence. The core task of pre-training is to train the model to recover these masked data points.

This innovative pre-training approach not only deepens the model's understanding and familiarity with the dataset but also highlights the novel value of this study. Through the process of striving to recover missing data points, the model thoroughly learns the inherent logic and structural features of the data, simultaneously enhancing its ability to handle incomplete or noisy data. This training method endows the model with greater stability and adaptability, making it suitable not only for regions with well-established infrastructure but also for resource-limited or economically underdeveloped areas. In these regions, despite potential damage or performance degradation of hardware devices due to various factors, our model can still maintain high prediction accuracy, providing reliable detection services. This aligns with one of our initial intentions in designing this model: to ensure stable and high-quality fetal monitoring services even in constrained conditions.

## Algorithm development

### Algorithm 1:

In this study, we propose a novel binary classification evaluation transformation strategy for time series analysis (see Fig. 2c). Given that acceleration, deceleration, and smooth periods often intertwine within the overall time series, we have carefully classified the entire time series into intervals based on different states.

Specifically, since signals appear in specific intervals, when an interval consistently exhibits acceleration signals, we explicitly label it as 1; for continuous intervals that do not exhibit special signals, we uniformly label them as 0, and this label persists until the signal changes again. The labeling process for deceleration intervals follows a similar logical principle. Through this strategy, we successfully simplify the originally complex and variable time series evaluation problem into a more intuitive binary classification evaluation problem (see Fig. 2e). Simultaneously, to further simplify the computation process without sacrificing overall accuracy, we select the time point with the highest probability of duration occurrence as the counting baseline. Only when the duration of the acceleration or deceleration state exceeds its corresponding dynamic counting threshold do we record and label it.

### Algorithm 2:

In addressing model prediction issues, we observed that time series prediction results often do not perfectly align with labeled tags. Therefore, when labeling prediction windows, we introduce the Intersection Overlapping Labels (IOL) evaluation criterion (see Fig. 2d). This criterion, based on the fine division of a single window, aims to achieve clear labeling of prediction windows. In practice, we first select the longest continuous prediction label within a single window and set a threshold (IOL threshold) such that when the proportion of the predicted continuous label in the window exceeds this threshold, and simultaneously satisfies the conditions of the predicted acceleration length being greater than 8 and the deceleration length being greater than 6, we then label the window.

### IOL thresholds

After defining the above two key metrics, we use the true labels as a baseline reference to conduct a detailed comparative analysis of the IOL ratios of the predicted labels within the corresponding true label intervals. To comprehensively evaluate the model's performance, we plotted specificity and sensitivity score curves for acceleration and deceleration binary classification evaluations in the test set under different IOL threshold conditions. Through our experimental comparisons, we found that when the IOL exceeds 0.5, i.e., when the proportion of the predicted label to the true label is more than 50%, the score curves for acceleration and deceleration in the test set reach an ideal balance point (see Fig. 2f). This important finding provides a strong reference for subsequent model optimization efforts and offers new ideas and methods for research in related fields.

## Future work

At the practical level of clinical application, although we have simulated scenarios that Doppler ultrasound may encounter in specific situations, our work still requires further deepening in the face of more intricate challenges posed by wearable devices in real-world environments. Through in-depth communication with gynecological experts, we have become aware that our model still suffers from issues of omission and misdiagnosis when processing Doppler ultrasound data contaminated by interference. To address this challenge, we will delve into decoupling techniques for signals with interference and artifacts, and couple randomly generated artifacts with Doppler ultrasound signal images as an effective means of data augmentation.

Furthermore, beyond the tasks of calculating FHR signal periodic variability and amplitude variability that the current model performs, we aim to integrate more fetal heart detection tasks into the model. This is intended to comprehensively enhance the model's overall performance and delve into the coexistence and mutual exclusivity relationships that exist among multiple fine-grained labels. This is where the advantage of our large-scale dataset lies, and it also constitutes the challenge we face.

It is worth mentioning that our proposed model and the method for judging acceleration and deceleration exhibit significant innovation. In future work, we will continuously optimize the model and actively collect and integrate more clinically representative features for fusion, with the aim of further enhancing the model's overall performance and accuracy through more extensive validation.

**Data availability**

The data supporting the plots in this article and other findings of the study are available at the following URL: https://drive.google.com/file/d/1Nm2knKS1goTG0yBnf9TeL562uKglASOM/view?usp=drive_link

**Code availability**

The network model and data processing code mentioned in this article are available at the following URL: https://drive.google.com/file/d/1zaCk628HfvENyAjGpGASJ7oHYL7j_Mrz/view?usp=drive_link